\documentclass[journal]{IEEEtran}
\usepackage{cite}
\usepackage{amsmath,amssymb,amsfonts}
\usepackage{algorithm}
\usepackage{algpseudocode} 
\usepackage{multirow}

\usepackage{hyperref}
\usepackage{color}
\usepackage{graphicx}
\usepackage{textcomp}
\usepackage{wrapfig}
\usepackage{algpseudocode}
\usepackage{placeins} 
\usepackage{amsmath}
\usepackage{amsfonts}
\usepackage{soul} 
\usepackage{array}
\usepackage{stfloats}

\usepackage{makecell}

\begin{document}

\title{An Event-based Fast Intensity Reconstruction Scheme for UAV Real-time Perception}
\author{ Xin Dong$^1$, Yiwei Zhang$^1$, Yangjie Cui, Jinwu Xiang, Daochun Li, and Zhan Tu, \IEEEmembership{Member, IEEE}, 
\thanks{This research received no specific grant from any funding agency in the public, commercial or not-for-profit sectors. (Xin Dong and YiWei Zhang contributed equally to this work.) (Corresponding author: Zhan Tu.) }
\thanks{Xin Dong is with Hangzhou International Innovation Institute, Beihang University, Hangzhou 311115, China (e-mail: xindong324@buaa.edu.cn). }
\thanks{Yiwei Zhang, Yangjie Cui, Jinwu Xiang, and Daochun Li are with the School of Aeronautic Science and Engineering, Beihang University, Beijing 100191, China (e-mail: zhangyiwei@buaa.edu.cn; cuiyangjie@buaa.edu.cn; xiangjw@buaa.edu.cn; lidc@buaa.edu.cn). }
\thanks{Zhan Tu is with the Institute of Unmanned System, Beihang University, Beijing 100191, China (e-mail: zhantu@buaa.edu.cn).}
\thanks{Jinwu Xiang is also with Tianmushan Laboratory, Xixi Octagon, Yuhang, Hangzhou 310023, China (e-mail: xiangjw@buaa.edu.cn).}
}

\maketitle

\begin{abstract}

Event cameras offer significant advantages, including a wide dynamic range, high temporal resolution, and immunity to motion blur, making them highly promising for addressing challenging visual conditions. 
Extracting and utilizing effective information from asynchronous event streams is essential for the onboard implementation of event cameras.
In this paper, we propose a stream-lined event-based intensity reconstruction scheme, event-based single integration (ESI), to address such implementation challenges. 
This method guarantees the portability of conventional frame-based vision methods to event-based scenarios and maintains the intrinsic advantages of event cameras.
The ESI approach reconstructs intensity images by performing a single integration of the event streams combined with an enhanced decay algorithm. Such a method enables real-time intensity reconstruction at a high frame rate (typically 100 FPS). Furthermore, the relatively low computation load of ESI fits onboard implementation suitably, such as in UAV-based visual tracking scenarios. 
Extensive experiments have been conducted to evaluate the performance comparison of ESI and state-of-the-art algorithms.
Compared to state-of-the-art algorithms, ESI demonstrates remarkable runtime efficiency improvements, superior reconstruction quality, and a high frame rate. As a result, ESI enhances UAV onboard perception significantly under visual adversary surroundings. In-flight tests, ESI demonstrates effective performance for UAV onboard visual tracking under extremely low illumination conditions (2–10 lux), whereas other comparative algorithms fail due to insufficient frame rate, poor image quality, or limited real-time performance.

\end{abstract}


\section{Introduction}
\label{sec:introduction}
\begin{figure}[!t]
\centerline{\includegraphics[width=\columnwidth]{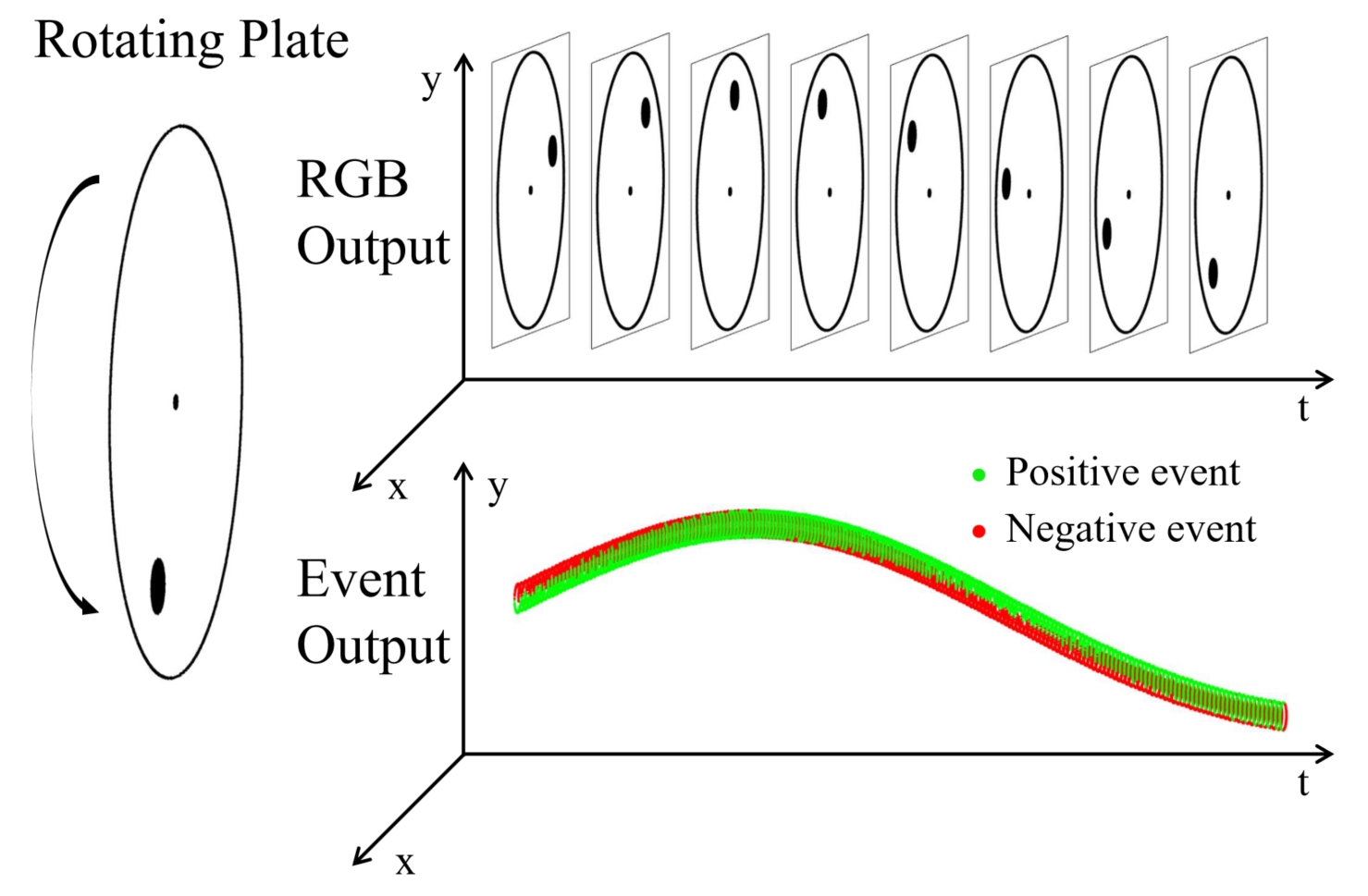}}
\caption{Comparison of conventional images and event streams. Conventional images are output at a fixed frame rate and contain intensity information for all pixels. The event stream is triggered by intensity changes and forms a temporally dense, spatially sparse event cloud.}
\label{fig1}
\end{figure}

\IEEEPARstart{C}{onventional} RGB cameras are widely used in unmanned aerial vehicle (UAV) onboard object detection\cite{zou2023object}, Simultaneous Localization And Mapping (SLAM)\cite{ebadi2023present}, and other visual perception domains\cite{rahman2020recent,yurtsever2020survey}. However, these cameras face inherent limitations such as motion blur, limited dynamic range, and low frame rate\cite{bao2019depth,zhang2022deep}. Under dim environments or high-speed motion, they often capture images of poor quality due to inadequate exposure and motion blur. These issues significantly weaken visual perception performance, which limits the operation of UAVs in such conditions accordingly. Considering these problems, an emerging bio-inspired dynamic vision sensor\cite{zou20171,son20174,suh20201280,lichtsteiner2008asynchronous,posch2010qvga}, namely, an event camera, offers a proper solution. Unlike the conventional camera, the event camera responds to the changes in log intensity and reports these changes as asynchronous pixel-wise measurements, known as event streams. By taking advantage of such a unique operating principle, the event camera boasts a wide dynamic range, high temporal resolution, and immunity to motion blur. Based on the above features, event cameras fit such platforms with real-time perception requirements, e.g., UAVs.
However, an event stream typically consists of millions of events per second, and its distinct format poses an open challenge to efficiently extract and utilize information\cite{gallego2020event}, as shown in Fig. \ref{fig1}.

Reconstruction intensity images are proposed to extract effective information from the event streams. 
Theoretically, the event streams encompass almost the entire intensity in the form of changes. Thus, it can be decompressed to reconstruct intensity images with the same favorable properties as the event camera\cite{zong2023single}. Moreover, the reconstruction process facilitates the portability of event cameras with conventional frame-based algorithms, thereby extending their applicability to all established domains of frame-based visual perception. Rebecq et al. show that conventional frame-based visual algorithms achieve better performance on the reconstructed image from the event streams for classification and visual-inertial odometry compared to dedicated event-based algorithms\cite{rebecq2019high}. This superior performance can be attributed to the mature development of conventional frame-based algorithms compared to event-based methods.

Current event-based intensity reconstruction schemes can be divided into two main categories: formation-model-based and learning-based intensity reconstruction. 
The former combines the event stream and conventional image to reconstruct intensity images without motion blur. However, as this category relies on the conventional image as the basis, the reconstruction frame rate is inherently limited. Typically, conventional cameras operate at 30 Frames Per Second (FPS), but the rate decreases in dim environments due to extended exposure time. As a result, the reconstruction frame rate can be significantly low, failing to meet the critical requirements of UAV real-time perception.
Learning-based methods have been employed to model the reconstruction process to eliminate reliance on conventional images. Typically, event streams are initially transformed into tensors, and then networks with various architectures are trained to directly reconstruct intensity images from these tensors. The complex structure of networks results in a significant increase in computational burden, making them unaffordable for UAV onboard applications.

To address the open challenge of UAV real-time perception under visually adverse conditions, in this work, we propose the Event-based Single Integration (ESI) method. Considering the high demands of real-time capability, ESI estimates the intensity value with a single integration operation. To further reduce computational load, denoising, and image mapping performance are fine-tuned. As a result, ESI enhances the algorithm for processing temporal event information, improving the adaptability to different scenes and boosting the quality of reconstructed images. The proposed ESI method was implemented onboard for UAV flight tests, which works effectively under dim conditions and during rapid motion, qualifying it for real-time perception.

The main contribution of this paper is that ESI is the first high-fidelity intensity reconstruction scheme capable of real-time, high-frame-rate (100 FPS) reconstruction on a computation-limited onboard computer. ESI enables real-time 100 FPS reconstruction on a TGU onboard computer. ESI demonstrates the proximate performance compared to the learning-based State of the Art (SOTA) methods. Note that such learning-based SOTA methods rely on RTX4070-based computation resources, so there are no examples of this being run on UAVs to date. In addition, we successfully conduct ESI-based onboard UAV flight tracking tests with 2 to 10 lux illumination. Based on flight test data, only ESI demonstrates sustained effective visual perception, whereas alternative algorithms fail due to reduced frame rate, degraded image quality, or limited real-time performance.

The remainder of this paper is organized as follows. Section \ref{sec:relate} reviews related work on event-based intensity reconstruction. Section \ref{sec:esi} details the proposed ESI scheme. Section \ref{sec:exp} presents the experimental results of both simulation and real-world experiments. Finally,  Section \ref{sec:conclu} concludes the remarks of our work.

\section{Related Work}
\label{sec:relate}

\subsection{Formation-Model-based Intensity Reconstruction}
Brandli et al. integrate the event streams to generate gray-scale images and incorporate a negative feedback mechanism to refine the estimated threshold\cite{brandli2014real}. However, this scheme is limited to stationary event cameras. Pan et al. propose a double integral model based on Brandli's work, applicable to moving event cameras\cite{pan2019bringing}. They define a cost function based on the sharpness and fitness of the edge to optimize the threshold. However, the complex optimization process hinders real-time implementation. Pan et al. extend the double integral model to the multiple event-based double integral model to reconstruct smoother intensity images from various images and events\cite{pan2020high}. Yet, the scheme is still based on a time-consuming optimization process and needs better real-time performance. Recently, Lin et al. improved the integration process of the double integral model and proposed the fast event-based double integral model (FEDI), enabling real-time operation for the first time \cite{lin2023fast}. The effectiveness of FEDI has been demonstrated through the application of reconstructed images to various algorithms, including SURF\cite{bay2006surf}, AprilTag\cite{garrido2014automatic}, and VINS\cite{qin2018vins}.

Scheerlinck et al. propose a complementary filter to fuse conventional images and event streams, which outputs continuous-time intensity estimation\cite{scheerlinck2018continuous}. This algorithm can also operate solely on event streams while reconstructing images with grayscale characteristics.
De et al. proposed a high-throughput asynchronous convolution method for image reconstruction by combining asynchronously accumulated events with temporal decay and sparse event-based convolution\cite{de2022high}. This method significantly improves processing throughput and is suitable for high-resolution event cameras, but the paper lacks quantitative evaluation of reconstruction quality.
The supporting software for the DAVIS series event camera provides a plugin called Accumulator\cite{ros2024dv}, which can reconstruct intensity images by accumulating the event streams. The plugin eliminates the reliance on conventional images, showing a high reconstruction frame rate. The major problem with the plugin is the severe trailing artifacts and shadows in the reconstructed images.

\subsection{Learning-based Intensity Reconstruction}
With the development of convolution neural network, recurrentneuralnetworkandgenerate
 against network deep learning model development, leveraging deep learning techniques to learn the mapping from event data to reconstructed images\cite{du2024overview,wang2021joint}. 
Rebecq et al. segment events into tensors along the temporal axis and process them through a recursive convolutional neural network E2VID to reconstruct intensity images\cite{rebecq2019high}. Although its training data comes from a simulation dataset, experiments demonstrate the generalizability to real-world scenarios. Wang et al. also utilize event tensors to train a conditional generative adversarial network, yielding reconstructed images with high frame rate and wide dynamic range\cite{wang2019event}. Scheerlinck et al. propose a novel neural network architecture to perform fast image reconstruction from events, named FireNet\cite{scheerlinck2020fast}. FireNet relies on the recurrent connection to build a state over time and reuses previous results, greatly reducing the parameter, memory, and FLOP demands compared to E2VID. However, it is still far from being able to operate in real-time on a single CPU. Paredes-Vallés et al.\cite{paredes2021back} introduced a self-supervised learning framework that utilizes photometric constancy to reconstruct images from event streams. This method eliminates the need for ground-truth images, making it more scalable and practical for real-world applications. More recently, Quan and Zhang\cite{quan2024image} provided a comprehensive survey of image reconstruction approaches based on the fusion of event streams and image frames, highlighting the potential of combining event data with traditional image data to enhance reconstruction quality. Additionally, Dauner\cite{dauner2023image} explored image reconstruction from event cameras specifically for autonomous driving applications, demonstrating the potential of learning-based methods in specialized domains. These learning-based approaches have shown significant improvements in reconstruction accuracy and robustness compared to traditional methods, thanks to their ability to learn complex patterns and relationships from large-scale data.

Despite the high reconstruction quality achieved by learning-based image reconstruction methods, their high computational loads  make them unacceptable for UAV onboard real-time application.

\section{Event-based Single Integration}
\label{sec:esi}

\subsection{Intensity Estimation}
Each pixel of the event camera responds to the changes in log intensity asynchronously. Once the change reaches the threshold \( c \), i.e., \(\left\| L(x, y, t) - L(x, y, t_k) \right\| \geq c \), an event will be triggered, denoted by \( e = \{x, y, t, p\} \). \( L \) denotes the log intensity, \( x \) and \( y \) represent the coordinates of the pixel, \( t \) is the time, and \( p = \{+1, -1\} \) denotes the polarity of the event, indicating brightening or darkening, respectively. By integrating the event streams, the log intensity at two moments can be correlated with the following equation:
\begin{equation}L(t_2) = L(t_1) \left[ c \int_{t_1}^{t_2} e(s) \, ds \right]\label{eq1}\end{equation}
where  \( e(s) \) represents the event whose timestamp equals $s$.

As seen from Eq. \ref{eq1}, without the absolute intensity as a basis, it is inadequate to accurately calculate the intensity solely based on the event streams. However, we find that the moving object triggers substantial variations in intensity, which suggests that estimating intensity solely with the event is feasible. Consequently, we assume that the baseline intensity across all pixels is uniform. Thereby, the intensity can be calculated by the simplified equation provided below:
\begin{equation}\tilde{L}(t) = c \sum_{s \in [0,t]} e(s)\label{eq2}\end{equation}
where \( \tilde{L}(t) \) represents the estimated intensity. 

\begin{figure*}[ht]
    \centering
    \includegraphics[width=\textwidth]{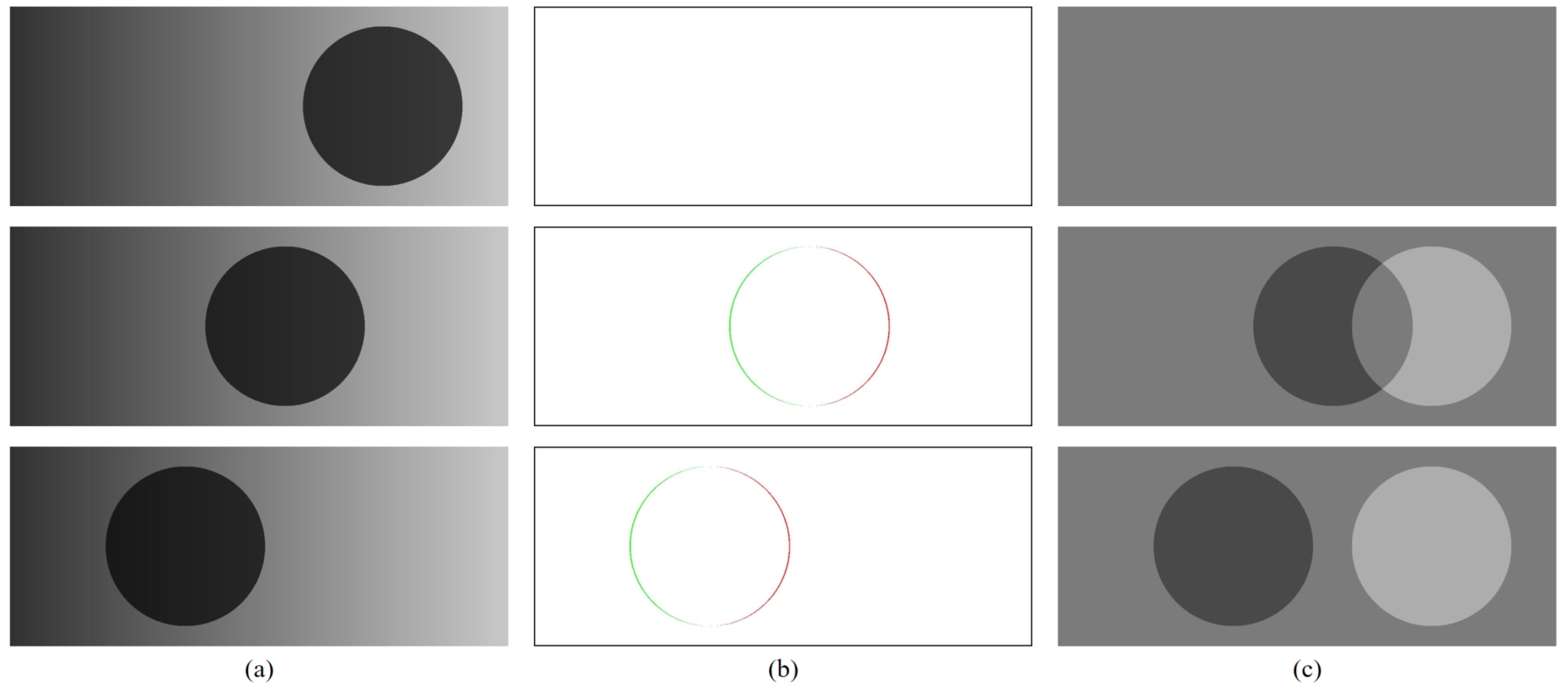}
    \caption{Simulation results. (a) ground truth intensity. (b) event visualization. (c) estimated intensity. In (b), red indicates positive events, while green indicates negative events.}
    \label{fig2}
\end{figure*}

We conduct simulation experiments on intensity reconstruction with uniform baseline intensity to validate this assumption. In the simulation, we set the raw intensity of the environment to increase linearly along the x-axis, with a circle with 30\% reflectivity moving along the negative direction of the x-axis, as shown in Fig. \ref{fig2} (a). The ground-truth intensity, visualization of the latest event stream, and the intensity estimated using Eq. \ref{eq2} are illustrated in Fig. \ref{fig2}. Initially, when the circle is stationary, no events are triggered, and the estimated intensity appears uniform due to the absence of baseline intensity difference. As the circle moves, positive and negative events are triggered on either side, leading to differences in the estimated intensity. In addition to the accurately represented circle, an erroneous reverse intensity estimation appears at the circle's initial position, which stems from the lack of baseline intensity difference. As the circle continues to move, the estimated intensity image still depicts the circle faithfully, yet the problem of reverse intensity estimation remains. This error will be addressed in the subsequent sections through the decay algorithm.

Overall, the intensity changes resulting from motion are significant, making it possible to ignore the baseline intensity difference and estimate intensity using the event streams only. The estimated intensity captures sufficiently distinct relative intensity relationships. Given that most visual perception algorithms rely on contrast, i.e., relative intensity instead of absolute values, to extract features and operate, the intensity estimated solely based on the event streams can adequately support visual perception.

\subsection{Decay Algorithm}
In the absence of baseline intensity difference, an erroneous reverse intensity estimation occurs at the initial position of the moving object. This problem is addressed through the decay algorithm discussed in this section. Moreover, the decay algorithm assists in resolving the severe noise associated with event cameras\cite{ding2023mlb}.
\begin{figure}[!t]
\centerline{\includegraphics[width=0.7\columnwidth]{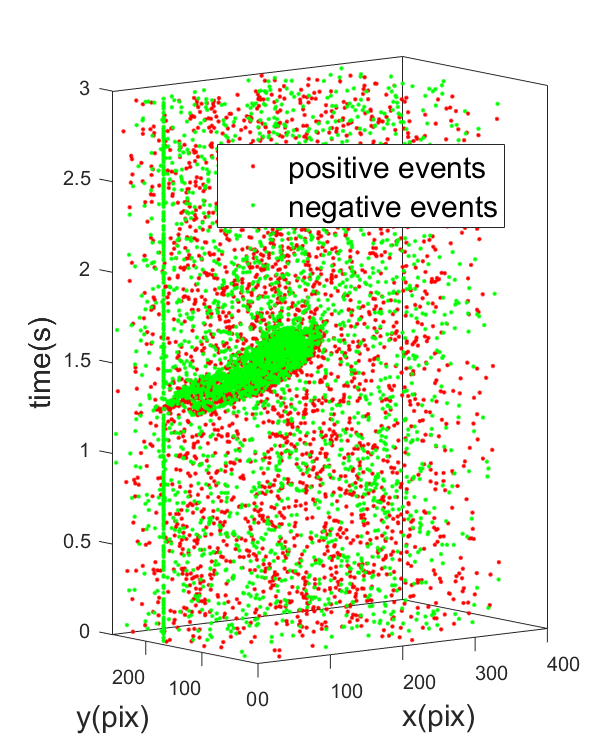}}
\caption{Visualisation of the real-world event streams. Events densely distributed in bands are triggered by the moving object, while scattered events indicate noise events, and the green straight line marks a hot pixel.}
\label{fig3}
\end{figure}

\begin{figure}[!t]
\centerline{\includegraphics[width=\columnwidth]{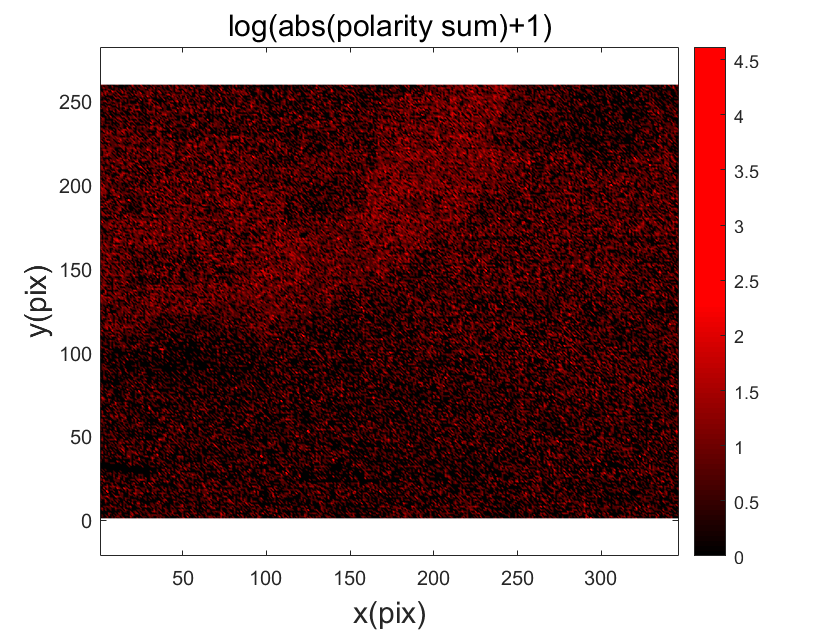}}
\caption{Visualization of noise accumulation.  The distribution and shade of red pixels illustrate how severe noise from the event camera can substantially degrade the estimated intensity. Noise-free pixels correspond to black, with deeper red indicating more severe noise accumulation.}
\label{fig4}
\end{figure}

A visualization of the captured event streams is shown in Fig. \ref{fig3} when an object is thrown before the event camera. For clear observation, only 1\% of the event streams are displayed. It can be observed that, in addition to the events correctly triggered by the moving object, a considerable number of noise events are scattered around. Accumulate the event streams from 0s to 2s with Eq. \ref{eq2}, take the absolute value, and add one before calculating the logarithm to get Fig. \ref{fig4}.

The intensities at 0s and 2s remain consistent, suggesting that the value in Fig. \ref{fig4} should be 0. Considering the threshold is 15\%, the value of 1.8 in Fig. \ref{fig4} represents a doubling in intensity, and 2.2 indicates a tripling. As shown in Fig. \ref{fig4}, the severe noise will quickly destroy the estimated intensity obtained from directly accumulating events. 

The decay function can be introduced to avoid the accumulation of noise events, which reduces the influence of events over time. With the introduction of the decay function, the intensity estimation function is modified as follows:

\begin{equation}\tilde{L}(t) = c \sum_{s \in [0,t]} e(s)d(t-s)
\label{eq3}\end{equation}
where \(d(t)\) represents the decay function. 

Once the object has moved away, no events are triggered at its initial position. Consequently, the inaccurately estimated intensity can be refined using the decay function. However, there are still two limitations that need to be addressed:

1. Memory Demands: Eq. ef {eq3} requires the storage of a considerable volume of events that still need to fully decay, thereby imposing substantial demands on memory resources. The event camera employed in this work is the DAVIS346 (resolution: 346 × 260 pixels), configured with a wide dynamic range of 120 dB and a temporal resolution of 1 $\mu \text{s}$. Under the bias and scene dynamics settings applied in our study, the sensor operates at a maximum output rate of 12 million events per second. Given this exceedingly high throughput, optimizing memory consumption becomes essential for ESI to meet the rigorous memory requirements of UAV onboard real-time perception.

2. Adaptability of Decay Algorithm: The decay algorithm of ESI should be adaptive to different scenarios without the need for fine-tuning parameters. Through extensive real-world experiments and statistical analysis, we find that the event-trigger rate is the primary feature distinguishing different scenarios. The decay should be slow to capture texture-rich areas in scenarios with high event-trigger rates. The estimated intensity closer to the ground truth through prolonged event retention enhances relative brightness relationships' representation. Conversely, regions with low event-trigger rates typically lack texture details and are predominantly composed of noise events. Consequently, these events should be decayed as soon as possible to reduce noise accumulation.

Considering the problems above, this paper proposes an improved decay algorithm to provide low memory demands and high environmental adaptability. The improved decay algorithm is specified in the pseudo-code below.

\begin{algorithm}
\caption{Decay algorithm}
\begin{algorithmic}[1]
\renewcommand{\thealgorithm}{}
    \State Initialize;
    \State $S(x, y) \gets 0$  //accumulation matrix;
    \State $T(x, y) \gets 0$ //last decay time matrix;
    \State \textbf{Input:} $k, b, C; E = \{e_1, e_2, ..., e_n\}$
    \For{each $e_i = (x_i, y_i, p_i, t_i)$ in $E$}
        \State //Update accumulation matrix $S(x, y)$
        \State $S(x, y) \gets S(x, y) + p_i \times C$
        \State //Calculate time difference $\Delta t$
        \State $\Delta t \gets t_i - T(x, y)$
        \State //Decay the accumulation value
        \State $S(x, y) \gets S(x, y) \times max\{(1 - k \times \Delta t)^b,0\}$
        \State //Update last decay time matrix $T(x, y)$
        \State $T(x, y) \gets t_i$
    \EndFor
    \State \textbf{Output final accumulation matrix $S(x, y)$}
    \State \textbf{return} $S(x, y)$
\end{algorithmic}
\end{algorithm}

In ESI, the decay function is defined as an exponentially weighted polynomial function, as shown in Eq. \ref{eq4}:

\begin{equation}
 d(t) = max\{(1 - kt)^b \, ,0\}, \quad k, \, b > 0,
\label{eq4}\end{equation}
where $k$ and $b$ represent the decay parameters.

The exponentially weighted polynomial function is highly adjustable and can model various decay curves. Moreover, ESI applies decay to the accumulation result. Therefore, the ESI only needs to store the accumulation result and the last decay time matrix for each pixel; this can greatly reduce the memory requirements. For a single pixel, the novel intensity estimation function can be described by:
\begin{equation}\ 
S(t) = c \sum_{i=1}^{n} \left[ e(t_i) \prod_{j=1}^{i} d(t_j - t_{j-1}) \right],
\end{equation}
where $n$ represents the index of the last event before $t$, satisfying \( t_n \leq t \) and \( t_{n+1} \textgreater t \).

Notably, the improved decay algorithm can automatically adapt to the varying event-trigger rate without tuning the parameter. In other words, compared to the conventional decay algorithm used in Eq. \ref{eq3}, ESI exhibits weaker decay effects when facing high event-trigger rates, thereby retaining more texture. This capability stems from the inequality as defined by the following equation:
\begin{equation}\ 
(d(t))^n > d(nt), \; n \in \mathbb{N}^+, \; n \geq 2 \;. \label{eq6}\end{equation}

It implies that the cumulative decay caused by multiple short intervals is less than that caused by a single long interval over the same period (the decay function describes the residual, where a higher residual means a weaker decay). The proof of Eq. \ref{eq6} is given below:

First, when $t < \frac{1}{nk}$ holds, Eq. \ref{eq6} can be expanded as:
\begin{equation}
(1 - kt)^{nb} > (1 - nkt)^b.
\label{eq7}
\end{equation}

Take the logarithm of Eq. \ref{eq7}, we have:
\begin{equation}
nb \ln (1 - kt) > b \ln (1 - nkt).
\label{eq8}
\end{equation}

Construct the function as:
\begin{equation}
F(t) = nb \ln (1 - kt) - b \ln (1 - nkt).
\label{eq9}
\end{equation}

Take the derivative to obtain:
\begin{equation}
F'(t) = nkb \left( \frac{1}{1 - nkt} - \frac{1}{1 - kt} \right).
\label{eq10}
\end{equation}

Obviously, for \( k, \ b > 0, \ n \in \mathbb{N}^+, \ n \geq 2 \text{ and } t \in \left(0, \frac{1}{nk}\right) \), it holds that:
\begin{equation}
F'(t) = \frac{n(n-1)k^2bt}{(1 - nkt)(1 - kt)} > 0.
\label{eq11}
\end{equation}

Combining \( F(0) = 0 \), there is:

\begin{equation}
F(t) > 0.
\label{eq12}
\end{equation}
The original inequality is proven.

\subsection{Clamping and Mapping}

As shown in Fig. \ref{fig3}, the green straight line corresponds to a hot pixel, which continuously outputs events. If the hot pixel is not suppressed, it will seriously inhibit other normal pixels. Consequently, it is necessary to perform clamping on $S(x, y)$ to prevent hot pixels or other pixels with abnormally high values from suppressing the rest. The clamping operation is described by the following equation:
\begin{equation}
S(x, y) = \min(\max(S(x, y), S_{\min}), S_{\max})
\label{eq13}
\end{equation}
where $S_{\max}$ and $S_{\min}$ represent the upper and lower bounds.

Clamping also improves the ESI's response speed to drastic changes in environmental illumination intensity. This improvement stems from the ability of clamping to directly filter out the influence of numerous events triggered by changes in environmental illumination intensity. In contrast, the decay function requires a long time to decay these events.

Finally, $S(x, y)$ will be mapped to an 8-bit gray-scale image. Typically, the mapping from intensity to a gray-scale image follows the logarithmic curve\cite{chou2013linear}. Since ESI accumulates events and each event corresponds to a change in log intensity, the linear mapping method can be employed, as described below:
\begin{equation}
B(x, y) = 255 \cdot \frac{S(x, y) - S_{\min}}{S_{\max} - S_{\min}}.
\label{eq14}
\end{equation}

The workflow of the proposed ESI is shown in the Algorithm \ref{alg:alg2}. The accumulation matrix and decay time matrix are initialized as zero. When the event streams are updated, the integration and decay are performed, and the intensity is estimated. Then, the clamping operation is applied to reject hot pixel outliers and increase robustness to drastic changes in environmental illumination. Finally, the intensity images can be obtained by converting the estimated value after clamping to an 8-bit grayscale image using a linear mapping method.

\begin{algorithm}
\caption{Event-based Single Integration}
\label{alg:alg2}
\begin{algorithmic}[1]
\State \textbf{Initialize}
\State $S(x, y) \gets 0$ // Initialize accumulation matrix
\State $T(x, y) \gets 0$ // Initialize last decay time matrix
\State \textbf{Input}
\State $k, b, C, E = \{e_1, e_2, ..., e_n\}$
\ForAll{$e_i = (x_i, y_i, p_i, t_i) \in E$}
    \State $S(x, y) \gets S(x, y) + p_i \times C$ 
    \State // Update accumulation matrix
    \State $\Delta t \gets t_i - T(x, y)$ 
    \State // Calculate time difference
    \State $S(x, y) \gets S(x, y) \times max\{(1 - k \times \Delta t)^b,0\} $
    \State // Decay accumulation value
    \State $T(x, y) \gets t_i$ 
    \State // Update last decay time matrix
    \State $S(x, y) \gets \min(\max(S(x, y), S_{\min}), S_{\max})$ 
    \State // Limit $S(x, y)$ to the range $[S_{\min}, S_{\max}]$
\EndFor
\For{$x = 1$ to $X$}
    \For{$y = 1$ to $Y$}
        \State $B(x, y) \gets 255 \times \frac{S(x, y) - S_{\min}}{S_{\max} - S_{\min}}$ 
        \State // Mapping $S(x, y)$ to an 8-bit gray-scale image
    \EndFor
\EndFor
\State \textbf{return} $B(x, y)$
\end{algorithmic}
\end{algorithm}

\section{Experiments}
\label{sec:exp}

In this section, we first evaluate the real-time performance of the ESI scheme by comparing the reconstruction time with the state-of-the-art algorithm. Subsequently, two experiments, object detection and visual tracking of UAVs, are conducted to demonstrate the advanced capabilities of ESI. A supplementary video can be found at \href{https://youtu.be/tLzXjXVRkVg}{https://youtu.be/tLzXjXVRkVg}.

\subsection{Reconstruction Runtime Evaluation}
\label{sec:expa}

To evaluate the reconstruction runtime, we compare ESI with alternative algorithms. For ESI, FEDI, the camera plugin, the complementary filter and HTAC, the experiment is performed with an Intel i7 CPU. As the two learning-based SOTA algorithms rely on GPU, they are tested on the PC with an RTX4070 GPU. The results are presented in Table. \ref{Table1}.

\begin{table}[t]
\centering
\label{table}
\caption{Runtime performance.}
\begin{tabular}{ccc}
\hline
\noalign{\smallskip}
Method               & \begin{tabular}[c]{@{}c@{}}Event process \\ capability(ev/s)\end{tabular}        & Platform                                                                  \\ \hline
ESI                  & 21.3*10\textasciicircum{}6  & \multicolumn{1}{c}{\multirow{4}{*}{\begin{tabular}[c]{@{}c@{}}Intel Core\\ i7-1185G7E@1.8GHz\end{tabular}}} \\
FEDI                 & 13.8*10\textasciicircum{}6  & \multicolumn{1}{c}{}                                                      \\
Camera plugin        & 25.4*10\textasciicircum{}6   & \multicolumn{1}{c}{}    
\\
Complementary Filter & 19.6*10\textasciicircum{}6  & \multicolumn{1}{c}{}                                                      \\ 
E2VID                & 0.40*10\textasciicircum{}6  & \multirow{2}{*}{RTX4070}                                            
\\  
FireNet              & 0.79*10\textasciicircum{}6 &     \\
\hline
\end{tabular}
\vspace{-0.1cm}
\label{Table1}
\end{table}

\begin{figure}[!t]
\centerline{\includegraphics[width=0.95\linewidth]{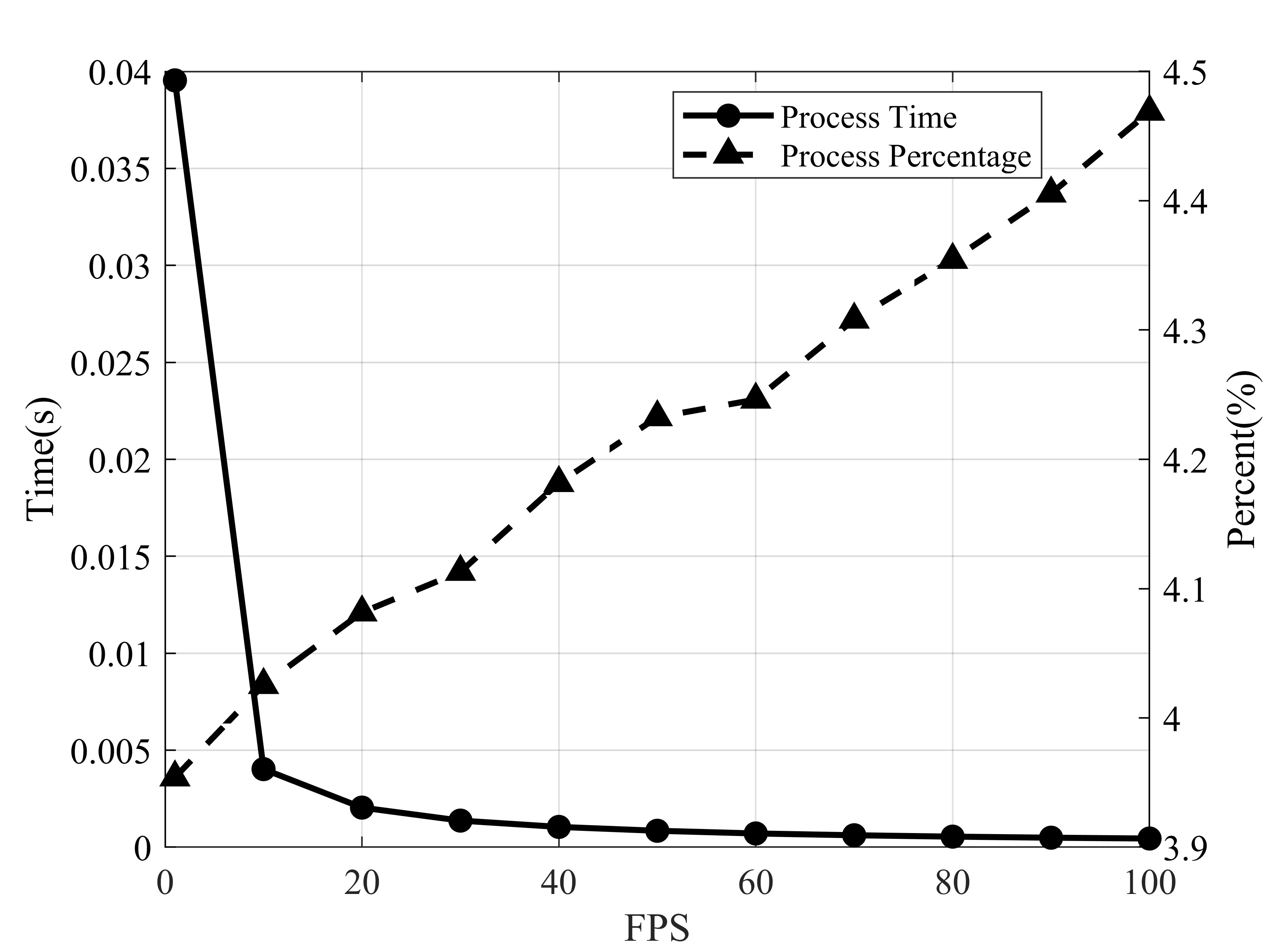}}
\caption{Resconstruction time per frame of ESI under different FPS settings.}
\label{fig:rec_time}
\end{figure}

ESI achieves high computational efficiency and is only slightly behind the camera plugin.
This is attributed to the increased computational complexity of the exponentially weighted polynomial function than the exponential function used in the camera plugin. 
The event camera used in this paper is DAVIS346 with an resolution of 346 $\times$ 260 pixels, a high temporal resolution of $1e^{-6}$s, with maximum output rate of 12 million events per second.

To further illustrate the computational efficiency of ESI, we have compared the processing time per frame and the percentage of frame processing time relative to the frame duration on our self-built dataset with different reconstruction frame rates (fps), the results as shown in Fig. \ref{fig:rec_time}. 

Therefore, ESI is capable of real-time operation using computationally limited resources even under extreme dynamic conditions.
Given that the DAVIS346 event camera packages asynchronous events and outputs them to the processor with a maximum frequency of 100 Hz, the reconstruction frame rate in the subsequent experiments is uniformly set to 100 FPS. 

\begin{figure}[!t]
\centerline{\includegraphics[width=\columnwidth]{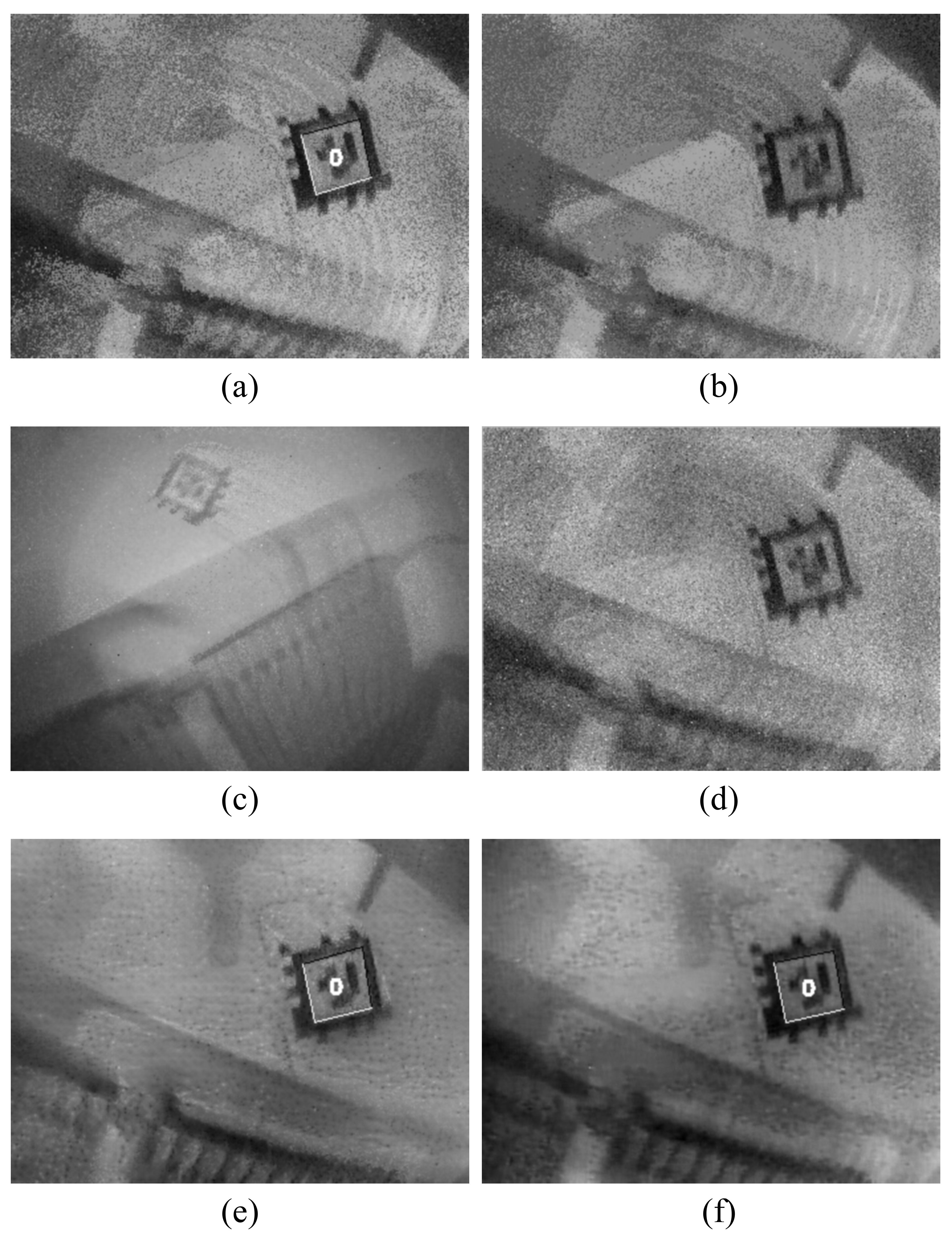}}
\caption{Comparisons of AprilTag detection. (a) ESI. (b) Camera plugin. (c) FEDI. (d) Complementary filter. (e) E2VID. (f) FireNet. White edge lines and the number “0” in the image indicate the successful detection, signifying high quality. These images show the detection results at the 21.62-th seconds in the experiment.}
\label{fig5}
\end{figure}

\subsection{Object Detection Performance Evaluation}
\label{sec:expb}

In this experiment, we employed the DAVIS346 event camera to capture the Apriltag marker. The data was first recorded by DAVIS346 camera as a rosbag package. Subsequently, the rosbag data was replayed and the images were reconstructed by proposed method and other algorithm to detect the marker. The experimental conditions include an illumination intensity ranging from 2 to 6 Lux, with the event camera moving around the stationary AprilTag. 
In this case, we implemented the proposed ESI, FEDI, the camera plugin and the complementary filter on the TGU8 onboard computer, and E2VID and FireNet on the PC with an RTX4070 GPU. The 2 to 6 Lux environment is notably dim, requiring the active pixels of the DAVIS346 to expose for 0.5 seconds to capture an adequately bright image. The experimental results are presented in Table. \ref{Table2}.

\begin{table}[t]
\centering
\label{table}
\caption{Results of AprilTag detection.}
\resizebox{\columnwidth}{!}
{
\renewcommand{\arraystretch}{1.3}
\begin{tabular}{ccccc}
\hline
\noalign{\smallskip}
Scheme & \begin{tabular}[c]{@{}c@{}}Frame rate\\ (FPS)\end{tabular} & \begin{tabular}[c]{@{}c@{}}Detection\\ count\end{tabular} & \begin{tabular}[c]{@{}c@{}}Detection\\rate\end{tabular} & \begin{tabular}[c]{@{}c@{}}Real-time\\capability\end{tabular}  \\ \hline
ESI                                                          & 100                                                        & 5046            & 81.39\%        & \multirow{4}{*}{\begin{tabular}[c]{@{}c@{}c@{}}Real-time\\ with TGU8 \\ onboard computer\end{tabular}   }                 \\
FEDI                                                         & 2                                                          & 18              & 14.52\%        &    \\
Camera plugin                                                & 100                                                        & 2871            & 46.31\%        &                                                                                      \\
Complementary filter & 100         & 2693               & 43.44\%         &                                                                                  \\
E2VID                                                        & 100                                                        & 5495            & 88.63\%        & \multirow{2}{*}{\begin{tabular}[c]{@{}c@{}}Post-process\\ with RTX4070\end{tabular}} \\
FireNet                                                      & 100                                                        & 5748            & 92.71\%        &                                                                                      \\

\hline

\label{Table2}
\end{tabular}}
\vspace{-0.5cm}
\label{Table2}
\end{table}

In experiment B, ESI, E2VID, FireNet, the camera plugin and the complementary filter reconstruct 6200 images at a preset frame rate of 100 FPS. The high frame rate should be attributed to all algorithms operating solely with the event streams, thereby fully leveraging the high temporal resolution of event cameras. In contrast, the exposure time of 0.5 seconds limits the frame rate of the conventional images to 2 FPS, capturing only 124 images. Due to its reliance on conventional images, FEDI reconstructs only 124 images as well.

Regarding runtime performance, ESI, FEDI, the camera plugin and the complementary filter can operate in real-time on the TGU8 onboard computer. As both E2VID and FireNet rely on the large-scale network for image reconstruction, even with an RTX 4070 GPU, they take 428.79 seconds and 255.73 seconds, respectively, to process the event streams of 62 seconds duration. Their runtime performance falls short of real-time operation for UAV onboard perception.

To better illustrate the quality of the reconstructed images, a comparative display is shown in Fig. \ref{fig5}.
It can be observed that the two learning-based SOTA schemes effectively extract spatial-temporal information from the event streams to mitigate noise and trailing artifacts, thereby achieving superior reconstruction quality. In contrast, ESI shows inadequate suppression of noise and trailing artifacts, leading to marginally lower image quality; however, this quality remains acceptable concerning detection rate.
Compared to the camera plugin and the complementary filter, ESI's improved decay algorithm more efficiently leverages the temporal information of events. It adapts to fluctuations in event-trigger rates, resulting in a 75.8\% and 87.4\% increase in detection rate, respectively. Their degradation is primarily reflected as the dark trailing shadows and low contrast, which impair the detection of the marker.
Furthermore, due to an extended exposure time of 0.5 seconds coupled with movement by the experimenter, conventional images suffer from significant motion blur and noticeable delays. Since FEDI relies on these low-quality conventional images as its foundation, reconstructing clear and sharp images becomes challenging, ultimately hindering the detection of Apriltag.

\begin{figure*}[htbp]
    \centering
    \includegraphics[width=\textwidth]{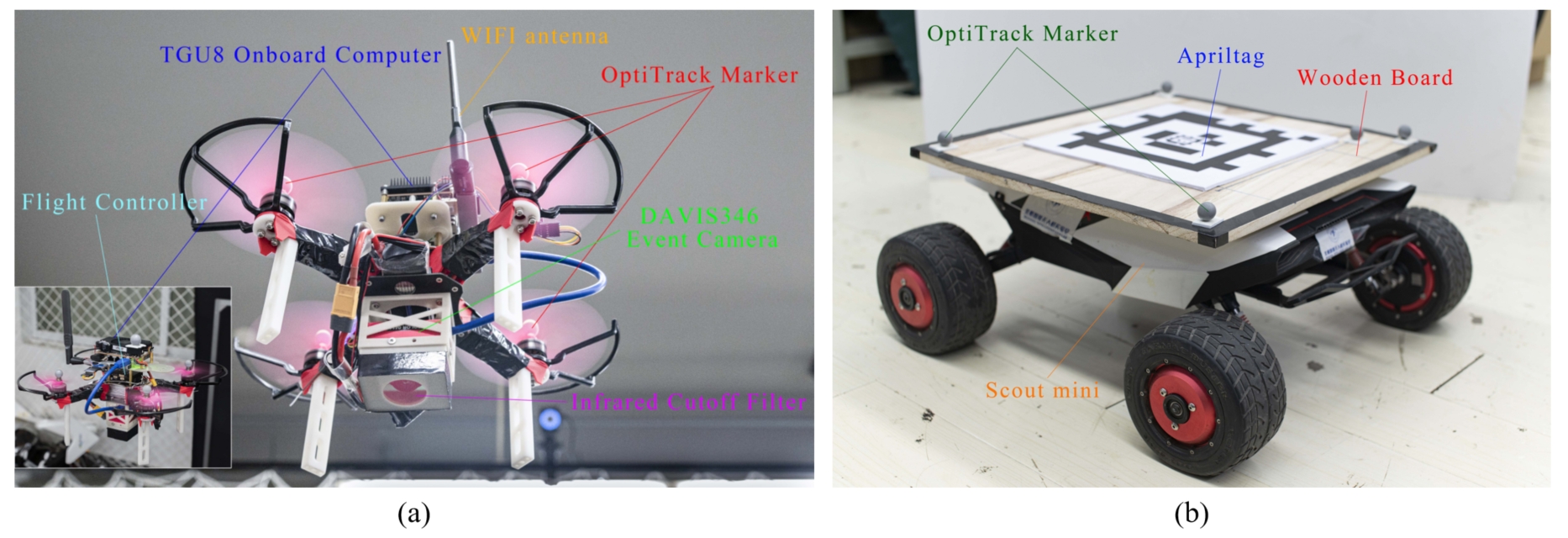}
    \caption{Experiment equipment. (a) UAV. (b) UGV.}
    \vspace{0.1cm}
    \label{fig6}
\end{figure*}

\begin{figure*}[htbp]
    \centering
    \includegraphics[width=\textwidth]{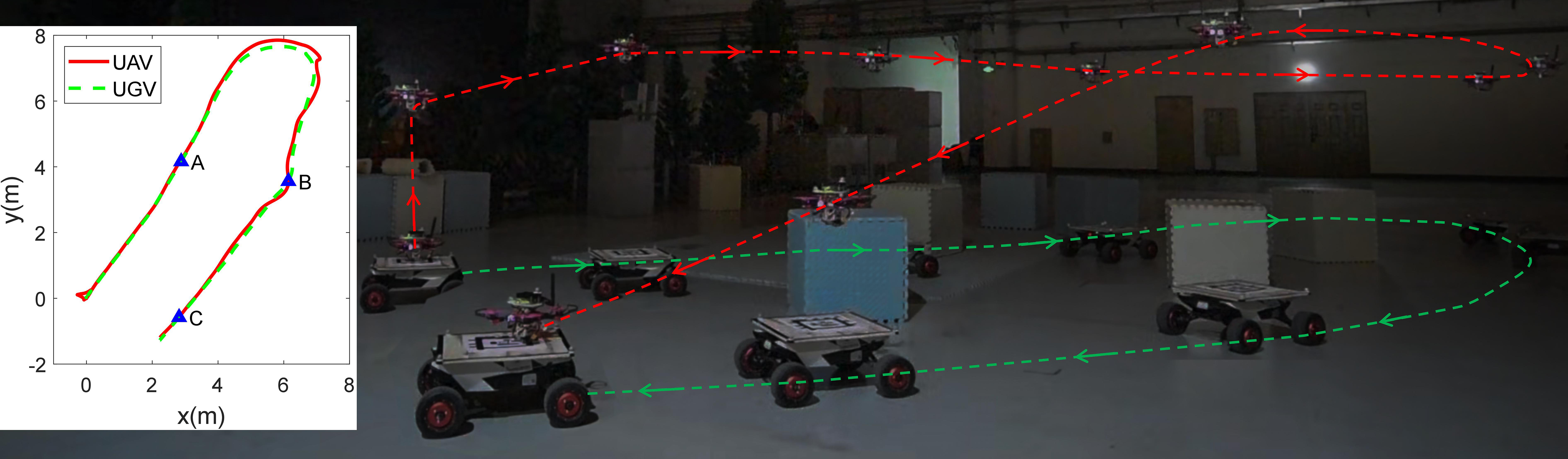}
    \caption{Trajectories of UAV and UGV.}
    \vspace{0.1cm}
    \label{fig7}
\end{figure*}

\begin{figure*}[htbp]
    \centering
    \includegraphics[width=\textwidth]{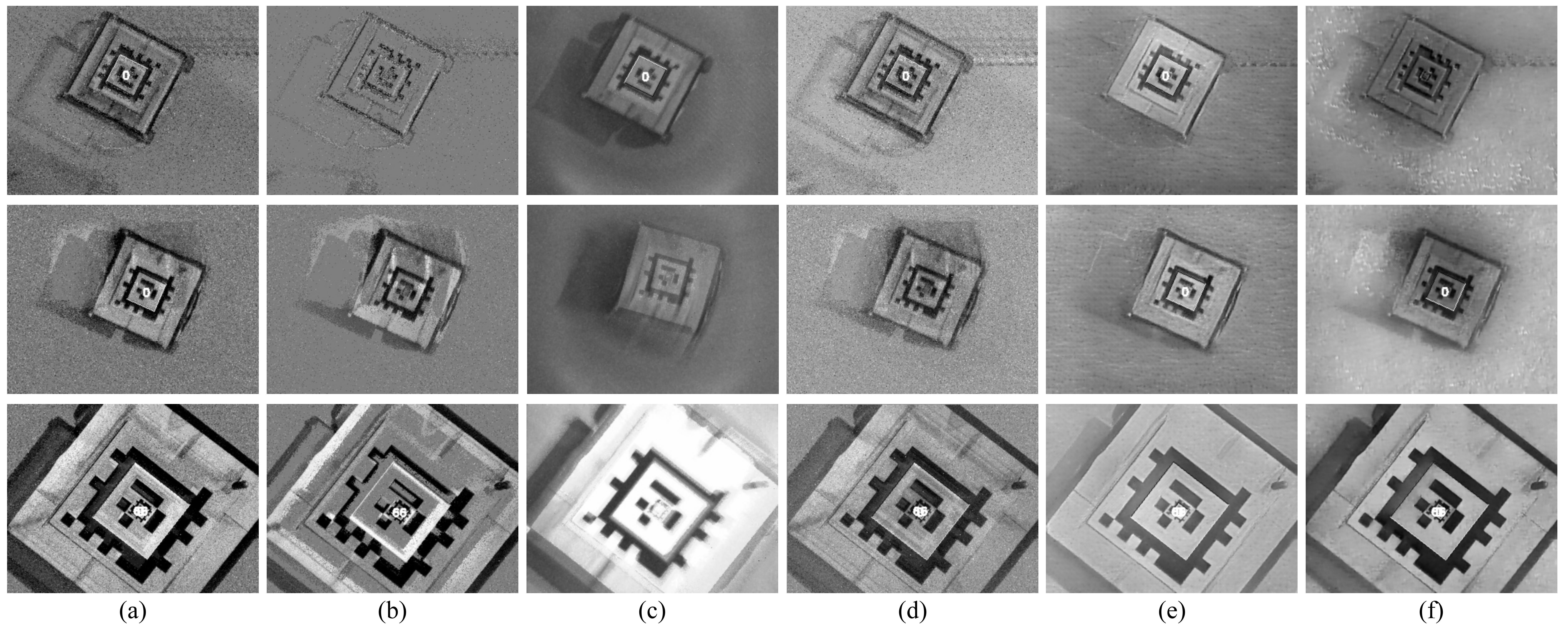}
    \caption{Comparisons of reconstructed images. (a) ESI. (b) Camera plugin. (c) FEDI. (d) Complementary filter. (e) E2VID. (f) FireNet. The three rows correspond to the trajectory points A, B, and C in Fig. \ref{fig7} from top to bottom.}
    \vspace{0.1cm}
    \label{fig8}
\end{figure*}

\subsection{UAV Visual Tracking Test}
\label{sec:expc}

To further verify the effectiveness of ESI in UAV onboard real-time perception, an experiment of UAV visual tracking is carried out in a dim environment with illumination intensity ranging from $2$ to $10$ lux. 

The UAV has a wheelbase of 250 mm and is equipped with a DAVIS346 event camera, a TGU8 onboard computer, and a Pixhawk 6cmini flight controller, while the unmanned ground vehicle (UGV) is equipped with the 30 cm $\times$ 30 cm AprilTag board, as shown in Fig. \ref{fig6}. The UAV locates the relative position by detecting the AprilTag from the reconstructed intensity images, and an extended Kalman filter is adapted to predict the motion state of the UGV. Image reconstruction, AprilTag detection, and relative pose estimation are all executed with the TGU8 onboard computer. The operating system is Ubuntu 20.04 with Robot Operating Systems (ROS), and the ROS driver of DAVIS camera\cite{mueggler2014event,lichtsteiner2008asynchronous,brandli2014spatiotemporal} is adapted to provide event messages for ESI. The ground truth position of both UAV and UGV is provided from the Optitrack motion capture system for validation. To prevent the infrared light of the motion capture system from affecting the event camera, an infrared cutoff filter is installed on the exterior of the event camera.

In the experiment, the UGV moves randomly by manual control, while the UAV employs the event camera to detect the AprilTag mounted on the UGV, estimates the relative pose, and tracks it automatically. Specifically, the UGV initially travels along a straight trajectory,  navigates through the obstacle area where it executes continuous turns to evade obstacles, and finally maintains its forward motion as the UAV lands on its wooden board. The trajectories of the UAV and UGV are depicted in Fig. \ref{fig7}, corresponding to the red and green dotted lines, respectively. 

The tracking process lasts for $62s$, and ESI provides $6540$ reconstructed images at $100$ FPS. The AprilTag is detected $4589$ times, achieving a success rate of $74\%$. Replaying the recorded data and reconstructing images through other schemes, the results are summarized in Table. \ref{Table3}. Reconstructed images of three typical phases of straight motion, turning and landing are selected to compare in detail, corresponding to points A, B, and C in Fig. \ref{fig8}, respectively.

As can be seen from Table. \ref{Table3} and Fig. \ref{fig8},  based on the detection results, ESI demonstrates a high frame rate and image quality.

\begin{table}[t]
\centering
\caption{Comparisons of detection counts.}
\resizebox{\linewidth}{!}{
\begin{tabular}{ccccc}
\hline
Scheme & \begin{tabular}[c]{@{}c@{}}Straight\\ motion\end{tabular} & Turning       & Landing      & Total          \\ \hline
ESI                                                          & \textbf{2626}                                                      & \textbf{1252}          & \textbf{711} & \textbf{4589}                                                         \\
FEDI                                                         & 51                                                        & 14            & 8            & 73                                                                        \\
Camera plugin                                                & 444                                                       & 493           & 154          & 1091                                                                      \\
Complementary filte & 2208 & 706 & 659 & 3573                                                                 \\

\hline
\end{tabular}
}
\vspace{-0.5cm}
\label{Table3}
\end{table}

The camera plugin suffers from pronounced shadowing and artifacts, which can be attributed to its mediocre decay function. This degradation becomes particularly evident during the straight motion phase, where the UAV tracks the UGV effectively, keeping both in a relatively static state and leading to a low event-trigger rate. The camera plugin suggests using a larger decay parameter to avoid noise accumulation, which results in severe image fading in such scenarios. Fortunately, the proposed ESI method introduces a certain level of adaptability. It allows ESI to effectively handle scenarios with varying event-trigger rates, ensuring high-quality reconstruction and successful detection.

As for the complementary filter, the adaptive decay parameter helps it partially adjust to the low-speed motion, resulting in a better detection performance compared to the camera plugin. However, during the turning phase, the limitations of the exponential decay cannot be fully compensated by the adaptive parameter, and the reconstructed images still exhibit a certain degree of shadowing. Ultimately, only an average detection performance can be achieved.

Regarding FEDI, its reliance on conventional images inherently limits its frame rate to 2 FPS, rendering real-time visual tracking impractical. During the straight motion phase, the UAV and UGV remain relatively static, and the quality of conventional images is high. Under this condition, both FEDI and raw conventional images enable successful detection, with FEDI producing higher-quality reconstructions through the event streams. However, during the turning and landing phases, conventional images are severely degraded by motion blur, preventing FEDI from producing evident reconstructions and leading to detection failures.

Overall, the proposed ESI scheme combines high frame rate, high image quality and low computational demands, enabling successful UAV visual tracking of the UGV in low-light environments. Notably, ESI achieves performance comparable to the two learning-based SOTA methods while significantly reducing computational power consumption, making it highly advantageous for UAV onboard real-time perception.

\section{Conclusion}
\label{sec:conclu}

This paper proposes the Event-based Single Integration (ESI) scheme to enable the event cameras' applications in UAV real-time visual perception under challenging conditions. 
ESI directly reconstructs intensity images from event streams, which enables the application of conventional frame-based visual algorithms, and avoids the complex computation of extracting visual information from asynchronous event streams. 
The reconstructed intensity images inherit the event camera's advantages of high dynamic range, high temporal resolution, and immunity to motion blur. Compared with the learning-based SOTA methods, ESI outperforms in runtime performance with proximate performance in reconstruction quality.
Various experiments are conducted to validate the reliability of the ESI algorithm in UAV perception applications under dim environments. 

\bibliographystyle{IEEEtran}
\bibliography{output.bib}

\vfill

\end{document}